\documentclass[preprint,5p,times,twocolumn]{elsarticle}

\usepackage{xcolor}
\usepackage{natbib}
\usepackage{hyperref}
\usepackage{amssymb}
\usepackage{url}
\usepackage{tabularx}
\usepackage{multirow}
\usepackage{array}
\usepackage{booktabs}
\usepackage{float}
\usepackage{xcolor}      
\usepackage{mdframed}    
\usepackage{amsmath}

\journal{Natural Language Processing}

\begin{document}

\newmdenv[
  linewidth=1pt,
  roundcorner=5pt,
  backgroundcolor=gray!5,
  linecolor=gray!75,
  innertopmargin=10pt,
  innerbottommargin=10pt,
  innerrightmargin=10pt,
  innerleftmargin=10pt,
  skipabove=12pt,
  skipbelow=12pt,
  frametitlealignment=\centering,
  frametitlefont=\bfseries,
  nobreak=true                 
]{myexamplebox}

\begin{frontmatter}



\title{It's All in The [MASK]: Simple Instruction-Tuning Enables BERT-like Masked Language Models As Generative Classifiers}
\author[label1]{Benjamin Clavié} 
\author[label2]{Nathan Cooper} 
\author[label2]{Benjamin Warner} 

\affiliation[label1]{organization={Answer.AI}, Japan}
\affiliation[label2]{organization={Answer.AI}, USA}

\begin{abstract}
While encoder-only models such as BERT and ModernBERT are ubiquitous in real-world NLP applications, their conventional reliance on task-specific classification heads can limit their applicability compared to decoder-based large language models (LLMs). In this work, we introduce ModernBERT-Large-Instruct, a 0.4B-parameter encoder model that leverages its masked language modeling (MLM) head for generative classification. Our approach employs an intentionally simple training loop and inference mechanism that requires no heavy pre-processing, heavily engineered prompting, or architectural modifications. ModernBERT-Large-Instruct exhibits strong zero-shot performance on both classification and knowledge-based tasks, outperforming similarly sized LLMs on MMLU and achieving 93\% of Llama3-1B’s MMLU performance with 60\% less parameters. We also demonstrate that, when fine-tuned, the generative approach using the MLM head matches or even surpasses traditional classification-head methods across diverse NLU tasks. This capability emerges specifically in models trained on contemporary, diverse data mixes, with models trained on lower volume, less-diverse data yielding considerably weaker performance. Although preliminary, these results demonstrate the potential of using the original generative masked language modeling head over traditional task-specific heads for downstream tasks. Our work suggests that further exploration into this area is warranted, highlighting many avenues for future improvements.
\end{abstract}

\begin{keyword}
Zero-shot classification \sep Multiple-choice question answering \sep Encoder models \sep BERT \sep ModernBERT \sep Masked Language Modelling \sep FLAN \sep Instruction-Tuning



\end{keyword}

\end{frontmatter}
\begin{figure*}[!htbp]
    \centering
    \includegraphics[width=0.8\linewidth]{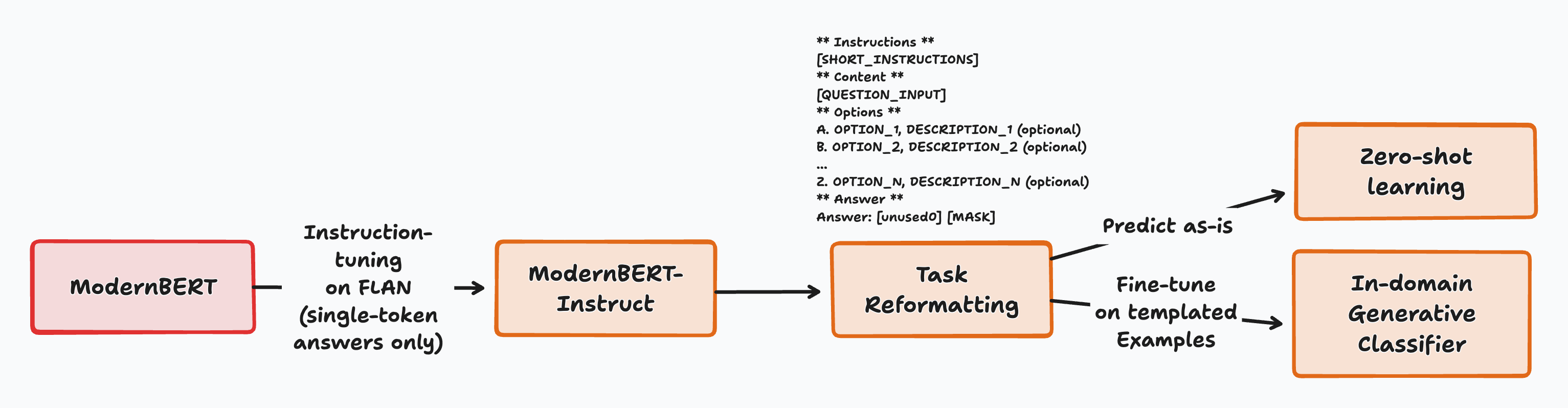}
    \caption{High level overview of the process.}
    \label{fig:overview}
\end{figure*}
\section{Introduction}

Since the release of BERT~\cite{bert} in 2018, Transformer~\cite{transformers}-based encoder-only pretrained models have been omnipresent in Natural Language Processing, being used to perform a wide variety of tasks and set state-of-the-art results across many of them. In the following years, much work has focused on further pretrained language models—with increasing emphasis on generative, decoder-only models trained on vast quantities of data, generally called Large Language Models (LLMs)—following the release of ChatGPT and the impressive performance of instruction-tuned AI assistants in general~\cite{tulu,llama3,qwen2.5}. Notably, LLMs have demonstrated their strong potential as zero-shot~\cite{gpt3} and few-shot~\cite{Radford2019} learners, especially when instruction-tuned~\cite{flan}.

However, while LLMs are very powerful, a large number of real-world tasks continue to be performed by encoder-only models, such as BERT, RoBERTa~\cite{roberta}, or DeBERTaV3~\cite{debertav3}. There are multiple reasons for the continued prevalence of encoder models. One important factor is their greatly reduced cost; they can process large volumes of data with computational requirements orders of magnitude lower than those of LLMs~\cite{modernbertbp}. Moreover, although decoder-only models are often better logical reasoners—thanks to their autoregressive nature and techniques such as Chain-of-Thought prompting~\cite{cot}—encoder-only models have consistently demonstrated strong performance on more traditional tasks, such as text classification, where fine-tuned encoder-only models can match the performance of LLMs that have 100× more parameters~\cite{clfsurvey}.

When using encoder-only models, classification tasks are traditionally performed with task-specific classification heads, where the model’s representations are pooled into a single vector that is then used to assign an appropriate label. However, this approach has traditionally lagged behind in zero-shot classification, as it requires a substantial fine-tuning step to yield good results. As a result, zero-shot classification methods have typically relied on leveraging the classification head by reframing tasks in an appropriate way. The dominant approach has been to repurpose textual entailment: after extensive pretraining on entailment tasks such as MNLI~\cite{MNLI}—where the model predicts whether a sentence is entailed or contradicted by another—the final model is used to determine whether a label is entailed by a given sentence~\cite{bart}. Further studies have shown that framing a very large number of tasks as entailment tasks can produce an even stronger zero-shot classifier~\cite{tasksource}. Another promising approach treats zero-shot classification as a form of next-sentence prediction~\cite{sst}.

Researchers have explored using the Masked Language Modeling (MLM) head—the generative component of encoder-only models—for classification. Early studies demonstrated strong potential in few-shot learning by turning tasks into Cloze-style questions~\cite{cloze}, where filling in a single token suffices to assign a label. For example, Pattern-Exploitive Training~\cite{PET} has been used to let models generate probabilities over multiple labels using multiple prompts, showing that these probabilities can transform MLM models into quasi-autoregressive few-shot learners~\cite{petfewshot} with only a few training steps per task. A recent study further demonstrated that MLM models can be converted into capable autoregressive in-context learners—requiring no additional training steps—by prepending multiple \textsc{[MASK]} tokens at the end of the document. However, this effect is only apparent in models with 900 million parameters or more~\cite{bertgenerative}.

Other studies have shown that prompt-based approaches can achieve strong results on classification and entailment tasks. However, these methods require custom prompts for each task and often perform worse than traditional classification-head methods~\cite{tasksource}. Moreover, these approaches are highly sensitive to the choice of tokens used to verbalize a label, with performance varying significantly even among semantically similar tokens. To address these issues, UniMC~\cite{unimc} introduced an additional step by converting tasks into a multiple-choice format using semantically neutral verbalizers (e.g., “A”, “B”) instead of meaningful words, and by employing a modified attention mask that ensures each verbalizer attends exclusively to its corresponding label description. Although this method achieved particularly strong results across a series of tasks, it adds considerable overhead.

Finally, it is worth noting that encoder-only models, while prevalent in the industry and representing the majority of model downloads from the HuggingFace Hub~\cite{modernbertbp}, are predominantly older models. In fact, the most widely used models are the original BERT~\cite{bert} (from 2018/2019), RoBERTa~\cite{roberta} (from 2020), and DeBERTaV3~\cite{debertav3} (from 2022), each with different strengths and weaknesses. This contrasts sharply with LLMs, where the vast majority of popular models are recent releases~\cite{llama3,qwen2.5,deepseek} and are updated multiple times per year~\cite{llama2,llama3,qwen2,qwen2.5}. These newer releases come with considerable improvements in many areas—particularly in the diversity and quality of data mixes—which have been shown to have a large impact on performance in zero- and few-shot tasks~\cite{dolma,dclm,fineweb}. The recent release of ModernBERT~\cite{modernbert}, trained using a modernized architecture and a large-scale, modern data mix, thus prompts further exploration into whether similar gains in zero-shot generative performance can be obtained with encoder-only models.

\subsection{Contributions}

Many existing approaches to generative classification using encoder-only masked language models rely on substantial overhead. These methods demand either extensive prompt design, require converting the model to operate in an autoregressive manner (thereby reducing the computational efficiency benefits of single-forward pass models), or necessitate adjustments to the attention mechanism.

In this paper, we showcase the potential of a simple generative approach to encoder-only models that requires little-to-no prompt engineering and no significant modifications to the model. We do so by introducing ModernBERT-Large-Instruct, an encoder-only model fine-tuned to perform classification tasks through its MLM head without further modification. We show that ModernBERT-Large-Instruct, although trained on the older FLAN instruction dataset, is remarkably strong for its size. Our approach is built around an extremely simple training loop to highlight its potential without heavy engineering requirements.

\textbf{Zero-Shot.} We show that ModernBERT-Large-Instruct is remarkably strong in multiple application settings. On classification tasks, it is competitive with previous approaches—outperforming all of them in true zero-shot settings on two out of three tasks. On zero-shot MMLU and MMLU-Pro, which evaluate the model’s knowledge and reasoning ability, it is the best overall model of its size class on MMLU, outperforming even similarly sized LLMs such as SmolLM2~\cite{smollm2} and performing closer to the much larger Llama3.2-1B~\cite{llama3}. On MMLU-Pro, it is the second-best model of its size class, losing only to another encoder model that employs an attention mask specifically designed to improve multiple-choice question performance.

\textbf{Fine-Tuning.} We also show that once finetuned, ModernBERT-Large-Instruct either outperforms or matches the performance of fully finetuned classification-head methods across a large number of common classification benchmarks spanning multiple domains—including news subject detection, textual entailment, forum post topic identification, and emotion detection—with even greater gains observed on more fine-grained tasks.

\textbf{Base Model Impact.} Finally, we demonstrate that these results are unique to ModernBERT~\cite{modernbert}. Models based on older architectures, such as RoBERTa-Large~\cite{roberta}, or those trained on less varied data mixes, such as GTE-en-MLM-Large~\cite{gte}, show noticeably worse performance. Our results join a growing body of research~\cite{unimc,PET,bertgenerative} and provide a strong basis for further exploration into leveraging the generative capabilities of masked language models. In particular, we believe that the effectiveness of our simple approach demonstrates the strong potential of low-parameter-count encoder-only masked language models.

\section{Methods}
\label{sec:training}

\subsection{Base Model}
\label{sec:basemodel}

We theorize that the ability to function as a generative multitask learner might be an emergent capability more modern, larger-scale data mixes, which have been shown to greatly improve zero-shot performance in decoder-only models~\cite{phi,dolma}. Another potential factor is the use of a more modern model architecture, more closely resembling the Transformer++~\cite{llama2,mamba} than the original Transformer~\cite{transformers}. Moreover, a model with a context length exceeding 512 tokens—the norm for most existing masked language models~\cite{bert,roberta}—allows for longer instructions and would therefore also be beneficial. We also believe that model size is a strong contributor in unlocking instruction-following, with Flan-T5 showing that substantial gains from multi-task instruction tuning are obtained at larger model sizes, with the smallest model variants showing much more modest improvements~\cite{flant5}.

We choose to use the recently released ModernBERT~\cite{modernbert}, which satisfies all of the above requirements, as it is an 8k context-length model using a modernized architecture and trained on a large-scale, modern data mix. We further explore this decision in Section~\ref{sec:backtothefuture} by comparing ModernBERT to other encoders, which do not satisfy all three constraints. Similarly to previous studies, we focus our explorations on ModernBERT-Large, the 395 million parameter ModernBERT variant, as is common practice in the exploration of encoder-only model zero-shot capabilities~\cite{unimc,sst}.

\subsection{Training Setting}
\label{sec:objective}

We propose a simple training objective to further pretrain a masked language model (MLM) to turn it into a capable zero-shot model. This training objective is a variant of the model's pretraining objective, utilizing the model's MLM head.

\textbf{Normal MLM} In classical MLM training, used to pretrain backbone models such as BERT~\cite{bert}, RoBERTa~\cite{roberta}, DeBERTa~\cite{deberta} or ModernBERT~\cite{modernbert}, a proportion of input tokens, between 15 and 40\%, is randomly masked: that is, replaced by a \textsc{[MASK]} token. The model's training objective then becomes a denoising objective: in a single pass, it must predict what the masked tokens originally were.

\textbf{Masked Instruction Tuning} The aim of this work is to functionally instruct-tune the Masked Language Modeling head of an MLM model to use its generative capabilities to perform a wide array of downstream tasks, in a way similar to sequence-to-sequence~\cite{seq2seq} models such as T5~\cite{t5} and its instruct-tuned variant, Flan-T5~\cite{flant5}. An immediate limitation of this method is that, unlike causal language modeling, masked language modeling generates its outputs in a single forward pass—replacing all \textsc{[MASK]} tokens simultaneously. Consequently, we must format our data so that the model is expected to predict only a single token for a given task.

\textbf{Answer Token Prediction} We propose a simple variant, inspired by large language models' instruction tuning pipeline: answer token prediction. This objective is a very simple tweak to the normal MLM objective. Rather than masking multiple tokens throughout the input text, we mask a single token, which is the \textbf{verbalizer} for a label or answer. This can be considered a restricted form of sequence-to-sequence learning\cite{seq2seq}, akin to the way generative models such as LLMs perform tasks. Effectively, this means that all tasks are reframed in the format of a \textbf{Cloze} question, where the input is formatted so answering the question requires generating a single verbalizer token. This templating process is further detailed in Section~\ref{sec:template}.

\textbf{Verbalizers} In the encoder-prompting literature~\cite{promptelectra}, a verbalizer refers to a single token that a model outputs to represent its entire answer. Such a token is considered semantically meaningful if it accurately describes the label (for example, the token “Positive” in the context of emotion classification). However, in many cases, it is not possible to express the correct model output with a single meaningful token. In these cases, semantically empty verbalizers, such as “A”, “B”, “C”, “D”, etc., are used to identify potential answers.

\subsection{Training Data}
\label{sec:data}

\begin{figure*}
    \centering
    \includegraphics[width=\linewidth]{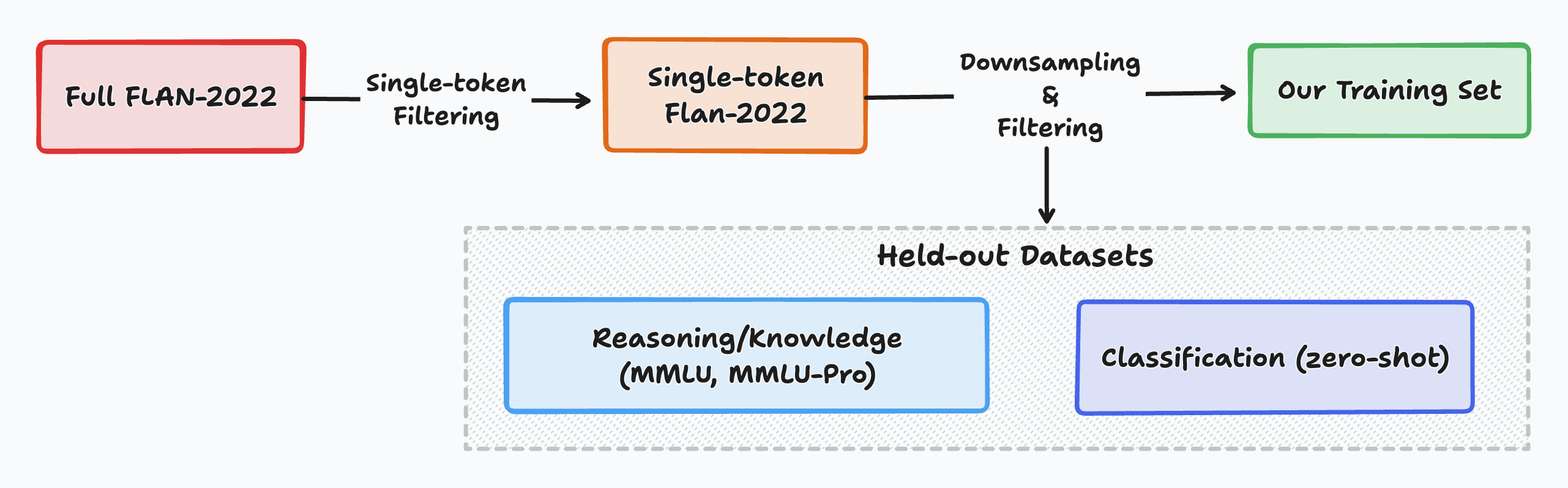}
    \caption{High level overview of our data processing pipeline.}
    \label{fig:data}
\end{figure*}

\textbf{The Single Token Constraint} Since MLM models are not pretrained in a generative fashion but rather to fill all \textsc{[MASK]} tokens, adding multiple \textsc{[MASK]} tokens could bias the model to consistently predict the longest possible answer. Rather than train the model to ignore extra masked tokens, a setting we leave for future work, we elected to reformulate the instruction tuning data and downstream tasks to require a single \textsc{[MASK]} token response.

\textbf{Modern Instruction Sets} This severely constrains our data selection process, as many of the newer instruction-tuning sets, such as T\"uluV3~\cite{tulu}, are explicitly constructed to train helpful instruct-tuned assistants. The expected model outputs are noticeably longer than a single token. As a result, many of the most popular instruction-tuning datasets are not suitable for this exploratory work, as they would require considerable pre-processing to be usable.

\textbf{FLAN} Older instruction-tuning datasets which predate the spread of assistant-style models ushered in by ChatGPT, such as the FLAN-2022~\cite{flan,flancol} collection, contain a large proportion of examples where the expected model answer is a single token. FLAN is a large-scale instruction-tuning dataset which contains hundreds of datasets further divided into thousands of tasks, with each task’s associated dataset preprocessed to match a specific template. This multi-task fine-tuning data has previously been shown to allow sequence-to-sequence models to reach improved downstream performance across all tasks, including previously unseen tasks~\cite{flant5}. 

\textbf{Filtering} The FLAN collection, in total, contains 396 million examples, of which 120 million are single-token answers. However, a large proportion of those examples come from just a handful of large datasets. The traditional approach to training on FLAN involves downsampling the data~\cite{flant5} to cap the number of examples from a single dataset. The FLAN authors originally filtered MMLU~\cite{mmlu} and BBH~\cite{bbh} from the data, to serve as held-out test sets. We follow their example and additionally filter out certain classification datasets which we use as downstream evaluation tasks, as detailed in Section~\ref{sec:classification}. A high-level overview of this filtering process is presented in Figure~\ref{fig:data}. Following the filtering and down-sampling step, our final training dataset contains \textbf{20 million examples}.

\subsection{The Elephant in The [MASK]: Dummy Examples Improve Performance}
\label{sec:dummy}

\textbf{Objective Mix} One of the worries with this approach is the potential for catastrophic forgetting due to the relative over-representation of verbalizers with no strong semantic meaning. To attempt to mitigate this effect, we experimented with introducing an Answer Token Prediction vs MLM ratio. Instead of applying answer token masking to all samples, we replace Answer Token Prediction on a subset of samples with standard MLM masking across all tokens. This results in each sample having either a single mask token for the answer, or thirty percent of all tokens randomly masked as during pretraining~\cite{modernbert}. Based on ablation runs, we choose a Answer Token Prediction vs MLM ratio of 20\%\footnote{15\% and 20\% MLM ratio performed similarly, with a slight edge for 20\%. Noticeable degradation was observed with ratios of 10\% and lower or 25\% and higher.}, meaning that 80\% of our examples used a single masked token objective, as described above, and the remaining examples used the typical random masking MLM pretraining objective.

\textbf{Dummy Examples} However, we introduced a labeling bug where all the true [MASK] tokens were incorrectly labeled as [MASK], rather than as their actual token. This meant that 20\% of the predicted tokens were of little value, as the model rapidly learned that in cases where multiple \textsc{[MASK]} tokens are present, their label is always the same, and the loss for these examples plummeted and stayed low throughout training. While we corrected this in a subsequent run, we decided to evaluate all three approaches: only answer token prediction, answer token prediction coupled with 20\% MLM, and answer token prediction coupled with 20\% "dummy MLM" variant.

\begin{table}[ht]
\centering
\begin{tabular}{lccc}
\toprule
Objective             & MMLU  & MMLU-Pro & Avg   \\
\midrule
ATP                   & 37.07 & 13.15    & 25.11 \\
0.8 ATP / 0.2 MLM     & 41.83 & 14.74    & 28.29 \\
0.8 ATP / 0.2 dummy   & \textbf{43.06} & \textbf{17.16} & \textbf{30.11} \\
\bottomrule
\end{tabular}
\caption{Comparison of MMLU results between the three training variants. 100\% ATP refers to using Answer Token Prediction as the only training objective. 80\% ATP/20\% MLM refers to using the ATP objective for 80\% of samples and the normal MLM objective for the remaining 20\%. 80\% ATP/20\% dummy refers to using the ATP objective for 80\% of samples and dummy MLM examples for the remaining 20\%.}
\label{tab:dummy}
\end{table}

\textbf{Improved Performance} We present the performance comparison results in Table~\ref{tab:dummy}. We were surprised to discover that the dummy MLM-enriched variant consistently and noticeably outperformed the other methods. To rule out randomness, we replicated this result across multiple seeds. We hypothesize that this effect is due to the dummy examples acting as a form of regularization, perhaps in a way similar to dropout~\cite{dropout}. We believe that this \textit{label dropout} mechanism warrants further investigation and leave it to future work as it is out of the scope of the current paper. 

\textbf{Final Objective Choice} Based on the improved performance we have observed, all pretraining experiments described in this paper use the training objective variant where 80\% of the training examples are answer token prediction. The remaining 20\% of training samples constitute dummy examples, where tokens are masked as they would be in normal MLM pretraining, but given a single, meaningless label.

\subsection{Templating}
\label{sec:template}

In order to leverage the Masked Language Modeling head for various tasks, we employ light templating. This templating process is different for training and inference. At training time, our goal is to maximize the number of template variations seen, while at inference time we use simple templates that require minimal modification for new tasks.

\subsubsection{Training}

During instruction training we add little additional templating to FLAN. Instead, we choose to reuse the extensive templating work already present in the FLAN dataset, based on the assumption that a diverse way of presenting similar tasks will lead to stronger generalization in the final model.

Accordingly, we apply a single templating step to all training examples: we prepend the \textsc{[MASK]} token (which represents the expected answer) with a previously untrained token that serves as an anchor~\footnote{We use the \textsc{[unused0]} token present in the ModernBERT vocabulary and left purposefully untrained to support this sort of application.}, indicating that the following token is the model’s final prediction. This is inspired by theof prefix tokens in the neural information retrieval literature to help the model distinguish between queries and corpus documents~\cite{colbert,e5,stella}.

\begin{myexamplebox}[frametitle={Example Training Instance}]
Frühere Versionen von KDE erlaubten mit einem Trick die Verwendung der Windows\textregistered-Taste
als Modifizierer (so dass Kurzbefehle wie \texttt{Win+R} möglich waren) und als reguläre Taste
(dass mit Betätigen der Taste \texttt{Win} das K-Menü geöffnet wurde).

Which language is this?
\textcolor{orange}{[unused0]} \textcolor{red}{[MASK]}
\end{myexamplebox}

Above is a randomly sample training example, with the \textsc{[unused0]} anchor token in orange and the masked answer in red.

\subsubsection{Inference}

During inference, we apply a simple formatting template, with minor tweaks depending on the task. The inference template follows a simple sequence, where (1) basic instructions are given to the model, followed by (2) the text to be labeled or responded to, (3) the list of potential labels, and finally, (4) an answer section that ends in a \textsc{[MASK]} token for the model to predict.

\begin{myexamplebox}[frametitle={Example Inference Instance (MMLU~\cite{mmlu})}]
You will be given a question as well as a list of options. Read the question carefully and select the right answer from the list.

QUESTION: \textcolor{blue}{An expected side effect of creatine supplementation is:}

CHOICES:

- A: \textcolor{blue}{muscle weakness.}

- B: \textcolor{blue}{gain in body mass.}

- C: \textcolor{blue}{muscle cramps.}

- D: \textcolor{blue}{loss of electrolytes.}

ANSWER:

Answer: \textcolor{orange}{[unused0]} \textcolor{red}{[MASK]}
\end{myexamplebox}

\section{Zero-Shot Performance}

\subsection{Setting}
\label{sec:classification}

\begin{table*}[!htbp]
\centering
\begin{tabular}{ll|cc|ccc|c}
 &  & MMLU & MMLU-Pro & ADEv2 & \multicolumn{1}{l}{NIS} & OSE & Average \\ \hline
\parbox[t]{3mm}{\multirow{6}{*}{\rotatebox[origin=c]{90}{0.3-0.5B}}}
 & Tasksource-NLI & 36.08 & 16.54 & \textit{65.17} & 58.72 & 21.11 & {\textit{39.52}} \\
 & RoBERTa-Large-SST & 31.3 & 13.63 & 43.61 & 75.0 & \textbf{40.67} & {40.84} \\
 & UniMC & 38.48 & \textbf{18.83} & 23.29 & 73.96 & 36.88 & {38.29} \\
 & ModernBERT-Large-Instruct & \textbf{43.06} & 17.16 & \textbf{53.31} & \textbf{85.53} & 20.62 & \textbf{43.94} \\
 & SmoLM2-360M & 35.8\textsuperscript{\textdagger} & 11.38\textsuperscript{$\ddagger$} & - & - & - & - \\
 & Qwen2-0.5B & 33.7\textsuperscript{\textdagger} & 15.93\textsuperscript{$\ddagger$} & - & - & - & - \\ \hline
 \parbox[t]{3mm}{\multirow{3}{*}{\rotatebox[origin=c]{90}{1B+}}}
 & Llama3.2-1B & 45.83 & 22.6\textsuperscript{$\ddagger$} & - & - & - & - \\
 & SmoLM2-1.7B & 48.44 & 18.31\textsuperscript{$\ddagger$} & - & - & - & - \\
 & Qwen2.5-1.5B & \textit{\textbf{59.67}} & \textit{\textbf{32.1}}\textsuperscript{$\ddagger$} & - & - & - & -
\end{tabular}
\caption{Result for the models across the zero-shot tasks. Results in \textbf{bold} are the best zero-shot results for the under-0.5B parameter model class. Results in \textit{italics} are results where the model is in-domain rather than zero-shot, reported for thoroughness. Results in \textbf{\textit{italicized bold}} are the best results in the 1B-and-above parameter model class. - indicates that the model was not evaluated on this task. Param count ranges are indicated on the left. Results marked with a $\dagger$ are directly reported from the results of Ben Allal et al.~\cite{smollm2} and results marked with a  $\ddagger$ are directly reported from the MMLU-Pro Leaderboard~\cite{mmlupro}.}
\label{tab:zeroshot}
\end{table*}

\textbf{Commonsense Reasoning and Knowledge} We use two standard LLM benchmarks to evaluate the knowledge and commonsense reasoning capabilities of ModernBERT-Large-Instruct: MMLU~\cite{mmlu} and MMLU-Pro~\cite{mmlupro} In MMLU, models are asked general knowledge questions across 57 subjects and must select the correct answer out of 4 possible options. To further evaluate the models' robustness on this task, we use MMLU-Pro, a smaller evaluation set designed to be more challenging. In MMLU-Pro, all questions were filtered to remove trivial knowledge tasks and favor reasoning-heavy questions. Additionally, rather than 4 options, each question has 10, lowering the impact of random guesses and assessing the ability to distinguish among multiple similar answers.

\textbf{Zero-Shot Classification} Evaluating zero-shot classifiers is surprisingly hard, as a large proportion of common benchmarks have been in use for many years, and data contamination during the post-training phase is considerably harder to avoid. The Real-world Annotated Few-shot Tasks (\textbf{RAFT}) benchmark~\cite{raft} contains multiple tasks which are part of FLAN and other commonly used datasets. Its full evaluation process relies on a held-out evaluation mechanism via an online platform which is currently offline. As part of our evaluation, we choose three of the RAFT tasks. Two of them are publicly available and commonly used outside the RAFT suite itself, while the final one is the NeurIPS Impact Statement (NIS) task, where the model must determine if an impact statement involves AI safety risks or not. For this task, we use the annotated training set as our evaluation set\footnote{This results in a limited evaluation set, of only 50 examples. However, for this task, the test set itself is only 150 examples; as a result, we believe this sample is an acceptable proxy and a better alternative to data contamination present in most of the other datasets.}. The publicly available datasets are ADEv2~\cite{ade}, a medical domain task where the model is given a sentence and must determine whether it is related to a negative reaction to a medication or not; and One Stop English~\cite{onestopenglish} (OSE), a general domain reading comprehension task where experts rewrote a single sentence into three levels of grammatical complexity: elementary, intermediate and advanced, and in which the model must accurately guess the complexity level. To the best of our knowledge, this setup ensures that only a single task is in-domain for the evaluated models.

\textbf{Baseline - Encoders} We compare ModernBERT-Large-Instruct against three types of baselines. For all tasks, we evaluate other encoder-based zero-shot methods, which are the most commonly used low-parameter-count methods for zero-shot applications. These baselines are UniMC~\cite{unimc} and Self-Supervised-Tuned RoBERTa-Large~\cite{sst}. We also report the best-performing entailment-based method, ModernBERT-Large-Tasksource-nli~\cite{tasksource}, as a final baseline, in order to represent this popular family of methods. However, it is worth noting that this model is in-domain on two out of the three classification tasks since zero-shot entailment-based methods are trained on large corpora, encompassing hundreds of tasks, including the vast majority of publicly available classification tasks. Neither method based on Tasksource~\cite{tasksource}—a comprehensive collection of tasks rephrased in a format compatible with entailment training—or other popular approaches such as ModernBERT-large-zeroshot-v2.0\footnote{https://huggingface.co/MoritzLaurer/ModernBERT-large-zeroshot-v2.0} are truly out-of-domain on common benchmarks.

\textbf{Baseline - LLMs} We also evaluate low-parameter-count Large Language Models on the commonsense knowledge and reasoning tasks, but exclude them from evaluation on classification tasks due to their overall poor instruction-following ability and their tendency to produce outputs that do not match the classification constraints. On MMLU and MMLU-Pro, we report two classes of LLM results. The first consists of the similar sized SmolLM2-360M~\cite{smollm2} and Qwen2-0.5B~\cite{qwen2.5} models, providing a direct comparison. The second consists of their larger variants, Llama3.2-1B~\cite{llama3}, SmolLM2-1.7B~\cite{smollm2} and Qwen2.5-1.5B~\cite{qwen2.5}, models that are almost four times larger than ModernBERT-Large-Instruct, to present upper bounds. Whenever possible, we report the results obtained by the authors of SmolLM2~\cite{smollm2} for MMLU. This covers all models except Qwen2.5-1.5B and Llama3.2-1B, which we obtained by using the Language Model Evaluation Harness~\cite{lmharness}. For MMLU-Pro, we report results from its official leaderboard\footnote{\url{https://huggingface.co/spaces/TIGER-Lab/MMLU-Pro}}.

\subsection{Results}

Results for the zero-shot experiments are presented in Table~\ref{tab:zeroshot}.

\textbf{MMLU} While non-causal models are often thought to memorize less information than their causal counterparts~\cite{bertgenerative}, our results demonstrate the opposite: on MMLU, ModernBERT-Large-Instruct outperforms all similarly sized models, including both decoder and encoder baselines, with its results being closer to a halfway point between models of its own size and models in the next size class, which have around 4 times more parameters.

\textbf{MMLU-Pro} While ModernBERT-Large-Instruct is the second-best performing on the harder MMLU-Pro, the best-performing approach at the sub-1B scale is also an encoder-only model, with UniMC showing remarkably strong performance. We theorize that UniMC outperforms ModernBERT-Large-Instruct on MMLU-Pro due to its custom attention masks, ensuring the model represents all labels individually rather than with noise from neighboring labels, while our simple approach forces the model to process the entire input at once with all tokens attending to each other. However, this confirms our overall hypothesis: encoder-only models' MLM heads appear particularly well suited to zero-shot tasks, even at small model sizes.

\textbf{Classification} On ADEv2 and NeurIPS Impact Statements, ModernBERT-Large-Instruct beats all other zero-shot approaches~\footnote{Tasksource-NLI's training set includes ADEv2, meaning it is fully in-domain on this task}, indicating strong performance across domains. However, on One Stop English, it noticeably lags behind two of the other methods, UniMC and Self-Supervised-Tuned (SST) RoBERTa. Both of these models leverage mechanisms not based on a typical classification head, with UniMC leveraging the model's generative head coupled with custom attention masks, and SST employing a next-sentence prediction objective.

\textbf{Overview} While ModernBERT-Large-Instruct reaches the best zero-shot performance across the evaluated tasks on average, in both classification and knowledge-based QA, it appears to have clear weaknesses, as discussed above. Noticeably, these  results point to a clear pattern indicating that various encoder-based methods have different strengths, as evidenced by the large swings in performance across datasets. Interestingly, UniMC~\cite{unimc}, another MLM-head based classification method, although relying on more complex mechanisms, strongly outperforms ModernBERT-Large-Instruct on two datasets, being the only model to do so, while being noticeably behind on three others. These results further reinforce the strong potential of using “non-traditional” classification heads for zero-shot tasks with encoder models, and warrant further research into the best ways to combine these methods.

\section{Full-Finetune: Can One Head Do It All?}

\begin{table*}[!htbp]
\centering
\begin{tabular}{l|l|crc|ccl|c}
                      & MNLI           & Yahoo!         & \multicolumn{1}{l}{20ng} & AGNews         & SST-2         & IMDB          & SST-5          & Average        \\ \cline{1-4} \cline{5-9} 
ModernBERT (cls head) & 90.8\textsuperscript{\textdagger}           & 77.75          & \textbf{73.96}           & \textbf{95.34} & \textbf{97.1}\textsuperscript{\textdagger} & 96.52         & 59.28          & 84.39          \\
ModernBERT-Large-Instruct   & \textbf{91.03} & \textbf{77.88} & \textbf{73.96}           & 95.24          & 96.22         & \textbf{97.2} & \textbf{61.13} & \textbf{84.67}
\end{tabular}
\caption{Results comparing our approach with a traditional classification head when fully fine-tuned on downstream NLU tasks. For SST-2 and MNLI, marked with a \textdagger, we report the results of Warner et al.~\cite{modernbert}.}
\label{tab:ftresults}
\end{table*}

\begin{table*}[!htbp]
\centering
\begin{tabular}{l|cc|cc|c}
Model Backbone   & MMLU           & MMLU-Pro       & ADEv2            & NIS           & Average \\ \hline
ModernBERT-Large & \textbf{43.06} & \textbf{17.16} & \textbf{53.31} & \textbf{85.53} & \textbf{47.8}   \\
GTE-en-MLM-Large & 36.69          & 14.47          & 20.26          & 79.48          & 35.44   \\
RoBERTa-Large    & 33.11          & 14.47          & 16.24          & 44.0          & 26.44  
\end{tabular}
\caption{Downstream results of using different backbone with the same instruction-tuning process.}
\label{tab:robertawalkedsomodernbertcouldfly}
\end{table*}

\subsection{Setting}
Having observed the strong zero-shot performance of ModernBERT-Large-Instruct, we now seek to explore whether this behavior holds true in fully fine-tuned settings. The common approach to encoder-model classification is to fine-tune a task-specific head, relying on pooling and a final layer projecting the pooled representation to a hidden dimension matching the number of potential labels.

To do so, we further finetune the ModernBERT-Large-Instruct model resulting from our experiments in section~\ref{sec:training} on a variety of downstream tasks. We use a simple fine-tuning mechanisms where all examples are trained on the single token prediction objective described in Section~\ref{sec:objective}\footnote{In our limited experiments, it appears that injecting dummy MLM examples, as presented in Section~\ref{sec:dummy}, helped to improve the pretrained model's generalization ability but does not improve performance on task-specific fine-tuning, however more exploration of this phenomenon remains warranted.}.

\textbf{Data} For this section, we evaluate the models on an array of widely used classification and Natural Language Understanding (NLU) tasks. We select 6 widely used~\cite{sst,unimc,clfsurvey} classification benchmarks, comprising of 3 topic classification detection and 3 emotion classification tasks. The three topic classification tasks are AGNews, where a news tagline must be classified into one of 4 categories, Yahoo! Answers Topic~\cite{yahoo}, a large-scale dataset where messages from the Yahoo Answer forum must be classified into one of ten topics, and 20newsgroup, where posts from online newsgroups must be classified into one of 20 forum categories. The emotion detection tasks are IMDB~\cite{imdb}, SST-2 and SST-5~\cite{sst2}, where movie reviews or extracts from such reviews must be classified into either positive or negative, in the case of IMDB and SST-2, or a finer-grained 5-level classification ranging from Very Negative to Very Positive for SST-5. We also evaluate the model on MNLI~\cite{MNLI}, an extremely common benchmark of NLU capabilities~\cite{wang-etal-2018-glue}, where the model is expected to predict the textual entailment between two sentences: that is, whether a hypothesis is entailed by, neutral toward, or contradicted by a given premise.

\textbf{Baseline} We compare a fully fine-tuned, in-domain, ModernBERT-Large-Instruct to a fully fine-tuned, in-domain, ModernBERT-Large, to provide the fairest comparison possible. The models are not evaluated against any external baseline, as the purpose of this task is to evaluate how well ModernBERT-Large-Instruct performs in this setting compared to the more common classification-head based approach. No extensive hyperparameter sweep is conducted; all results reported are the results of a minimal sweep across epoch counts \textsc{[1,2,3]} and learning rates [2e-5, 3e-5, 5e-5] for the MLM setting and [3e-5, 5e-5, 8e-5] for the traditional classification approach. No prompt engineering is performed to keep the comparison as fair as possible.

\subsection{Result} 

The results are presented in Table~\ref{tab:ftresults}. Overall, it appears that with full-dataset fine-tuning, ModernBERT-Large-Instruct outperforms a fully fine-tuned ModernBERT-Large using a classification head on average. On a task-by-task basis, both approaches trade blows, with both methods coming slightly ahead on certain datasets. Given our extremely simple training loop and limited optimization, we believe that this highlights the robust performance and strong potential of MLM-head based downstream tasks when compared to classification-head based approaches.

While the traditional classification method performs better on simpler tasks, such as the SST-2 binary sentiment classification dataset, ModernBERT-Large-Instruct achieves noticeably better results on its more fine-grained version, SST-5, where the emotion expressed in a sentence is broken down into five levels rather than just two. ModernBERT-Large-Instruct's MNLI performance also highlights that this method is not restricted to single-sentence tasks, but is able to correctly label the relationship between two sentences.

\section{Are Older Encoders Also Generative Classifiers?}
\label{sec:backtothefuture}

A final question we seek to answer, raised in Section~\ref{sec:basemodel}, is whether or not this behavior is specific to ModernBERT, and thus is emergent to the combination of a more modern architecture along with a sizeable mixed-sources 2 billion token pretraining, or was already present in previous encoder models.

\subsection{Setting}

To answer this question, we reproduce the training setting method presented in Section~\ref{sec:training} on two other encoder-only models. The first model is RoBERTa-Large~\cite{roberta}, one of the best performing and currently the most widely used MLM model~\footnote{as judged by downloads on the HuggingFace hub}. RoBERTa-base falls within the category of “classical” encoder-only pretrained models, using an architecture very similar to the original BERT~\cite{bert} model, and trained on limited data sources. We also evaluate GTE-en-MLM-Large~\cite{gte}, which provides an interesting comparison: while it has been trained on a limited data mix, its architecture is modernized and follows the more modern Transformer++ architecture~\cite{llama2,mamba}.

\subsection{Results}

We present the outcome of this experiment in Table~\ref{tab:robertawalkedsomodernbertcouldfly}. The results are stark: RoBERTa-Large, despite a similar parameter count to ModernBERT-Large and almost competitive performance in traditional, classification-head based approaches, achieves poor results in all zero-shot contexts when using the MLM head. GTE-en-MLM-Large~\cite{gte}, with its modern architecture and longer context length, fares better than RoBERTa-Large on all tasks except MMLU-Pro, where they are equal, but remains considerably weaker than ModernBERT-Large-Instruct.

As such, we hypothesize that strong generalization potential from a Masked Language Model's MLM head is a property relying mainly on a large-scale, varied pretraining data mix, with architecture itself playing a smaller, but still important, overall role.

\section{Conclusion}

In this work, we explored the potential of encoder-only masked language models (MLMs) as competitive zero-shot and fine-tuned classifiers through their generative MLM heads. We achieve this by designing a very straightforward training recipe and introducing ModernBERT-Large-Instruct. This model is trained on a filtered subset of the FLAN dataset with a simple training objective. Interestingly, we observed that the use of an unconventional training setup, where 20\% of the examples are replaced by dummy samples, vastly outperforms more traditional training approaches.

We demonstrated that modern encoder models can achieve remarkably strong classification performance via the use of their masked language modeling head, in a variety of settings. In zero-shot contexts, ModernBERT-Large-Instruct is competitive with previous approaches requiring more complex training or architectural modifications, even outperforming them on some tasks. On zero-shot knowledge benchmarks such as MMLU, it performs closer to the considerably larger Llama-3.2-1B than the similarly sized SmolLM2-360M.

When fine-tuned for a single specific tasks, this language-modeling-head classification approach reaches stronger overall performance than traditional classification-head fine-tuning on a wide range of downstream domains, encompassing entailment, topic classification, and emotion detection. These early findings suggest that generative classification with fine-tuned encoder models might represent a better alternative to more commonplace classification approaches.

Finally, we show that these capabilities appear to be unique to encoder-only models trained with modern training recipes, requiring both a modern architecture and modern data mixes, such as ModernBERT. Both older models (e.g., RoBERTa) and modern models trained on lower volume, less-diverse data (e.g. GTE-en-MLM) showed significantly weaker performance across all tasks.

While these are early results, our findings join a small but growing body of work showing that encoder-only models can function well as generative multi-task models. We believe that our work, relying on very simple training and inference mechanisms with a model with a very small parameter count by modern standards (0.4B), demonstrates the strong potential of such approaches.

\section{Limitations \& Future Work}

\textbf{Newer Instruction Sets} Our work focuses on the exploration of FLAN. While FLAN remains a well-constructed dataset with strong previous empirical results, it is an earlier form of instruction tuning. In modern LLM instruction-tuning, FLAN has often been superseded by much broader instruction sets focusing on more diverse tasks and phrasings~\cite{tulu,openorca,hermes}, although many continue to include FLAN as a subset~\cite{tulu,openorca}. However, we choose to use FLAN as the more complex data pre-processing methods required to convert modern turn-based instruction datasets into single token data are out-of-scope for this study. As such, FLAN and its preprocessed format making adapting a large subset of it to a single token generation format very straightforward, represents an ideal experimentation dataset.

\textbf{Few-Shot and In-Context Learning} We focus on exploring the potential of ModernBERT-FLAN and, more generally, of leveraging the MLM head for classification tasks in both zero-shot and full fine-tuning contexts, showcasing its superiority over previous methods. However, an area of particular interest is few-shot learning, either via in-context learning~\cite{icl,bertgenerative} or sample-efficient fine-tuning~\cite{petfewshot}. We leave this to future work.

\textbf{QA-Specific Attention Masks} In this work, we focus on a simple training loop to highlight the baseline capabilities of ModernBERT-Large-Instruct. However, previous approaches relying on carefully crafted attention masks such as UniMC~\cite{unimc} remain very competitive, despite using an older backbone. We believe that our results should encourage further exploration into modernizing UniMC-like approaches, and we intend to explore this in future work.

\textbf{Single-Token Generation} As our approach relies on generating a single token at a time, this imposes huge constraints on model outputs and necessitates the use of single-token verbalizers, which might often be tokens that themselves carry limited semantic information, such as single letters or digits. We believe that future research should explore approaches such as the one highlighted by~\cite{bertgenerative} to determine whether such methods could further increase performance.

\textbf{Scaling} Our study focuses on a model with 395M parameters, while the existing instruction-tuning literature has shown that larger models benefit even more from such post-training, especially in the case of FLAN~\cite{ul2}. However, at the time of our work, no 1B+ masked language model with a recent training mixture was available. We believe that, should such a model become available, our method would transfer directly and yield even greater results.

\bibliographystyle{elsarticle-num}

\bibliography{biblio}

\begin{thebibliography}{10}
\expandafter\ifx\csname url\endcsname\relax
  \def\url#1{\texttt{#1}}\fi
\expandafter\ifx\csname urlprefix\endcsname\relax\def\urlprefix{URL }\fi
\expandafter\ifx\csname href\endcsname\relax
  \def\href#1#2{#2} \def\path#1{#1}\fi

\bibitem{bert}
J.~Devlin, M.~Chang, K.~Lee, K.~Toutanova, \href{https://doi.org/10.18653/v1/n19-1423}{{BERT:} pre-training of deep bidirectional transformers for language understanding}, in: J.~Burstein, C.~Doran, T.~Solorio (Eds.), Proceedings of the 2019 Conference of the North American Chapter of the Association for Computational Linguistics: Human Language Technologies, {NAACL-HLT} 2019, Minneapolis, MN, USA, June 2-7, 2019, Volume 1 (Long and Short Papers), Association for Computational Linguistics, 2019, pp. 4171--4186.
\newblock \href {https://doi.org/10.18653/V1/N19-1423} {\path{doi:10.18653/V1/N19-1423}}.
\newline\urlprefix\url{https://doi.org/10.18653/v1/n19-1423}

\bibitem{transformers}
A.~Vaswani, N.~Shazeer, N.~Parmar, J.~Uszkoreit, L.~Jones, A.~N. Gomez, L.~Kaiser, I.~Polosukhin, \href{https://proceedings.neurips.cc/paper/2017/hash/3f5ee243547dee91fbd053c1c4a845aa-Abstract.html}{Attention is all you need}, in: I.~Guyon, U.~von Luxburg, S.~Bengio, H.~M. Wallach, R.~Fergus, S.~V.~N. Vishwanathan, R.~Garnett (Eds.), Advances in Neural Information Processing Systems 30: Annual Conference on Neural Information Processing Systems 2017, December 4-9, 2017, Long Beach, CA, {USA}, 2017, pp. 5998--6008.
\newline\urlprefix\url{https://proceedings.neurips.cc/paper/2017/hash/3f5ee243547dee91fbd053c1c4a845aa-Abstract.html}

\bibitem{tulu}
N.~Lambert, J.~Morrison, V.~Pyatkin, S.~Huang, H.~Ivison, F.~Brahman, L.~J.~V. Miranda, A.~Liu, N.~Dziri, S.~Lyu, et~al., T$\backslash$" ulu 3: Pushing frontiers in open language model post-training, arXiv preprint arXiv:2411.15124 (2024).

\bibitem{llama3}
A.~Dubey, A.~Jauhri, A.~Pandey, A.~Kadian, A.~Al-Dahle, A.~Letman, A.~Mathur, A.~Schelten, A.~Yang, A.~Fan, et~al., The llama 3 herd of models, arXiv preprint arXiv:2407.21783 (2024).

\bibitem{qwen2.5}
QwenTeam, A.~Yang, B.~Yang, B.~Zhang, B.~Hui, B.~Zheng, B.~Yu, C.~Li, D.~Liu, F.~Huang, H.~Wei, H.~Lin, J.~Yang, J.~Tu, J.~Zhang, J.~Yang, J.~Yang, J.~Zhou, J.~Lin, K.~Dang, K.~Lu, K.~Bao, K.~Yang, L.~Yu, M.~Li, M.~Xue, P.~Zhang, Q.~Zhu, R.~Men, R.~Lin, T.~Li, T.~Tang, T.~Xia, X.~Ren, X.~Ren, Y.~Fan, Y.~Su, Y.~Zhang, Y.~Wan, Y.~Liu, Z.~Cui, Z.~Zhang, Z.~Qiu, \href{https://arxiv.org/abs/2412.15115}{Qwen2.5 technical report} (2025).
\newblock \href {http://arxiv.org/abs/2412.15115} {\path{arXiv:2412.15115}}.
\newline\urlprefix\url{https://arxiv.org/abs/2412.15115}

\bibitem{gpt3}
T.~B. Brown, B.~Mann, N.~Ryder, M.~Subbiah, J.~Kaplan, P.~Dhariwal, A.~Neelakantan, P.~Shyam, G.~Sastry, A.~Askell, S.~Agarwal, A.~Herbert{-}Voss, G.~Krueger, T.~Henighan, R.~Child, A.~Ramesh, D.~M. Ziegler, J.~Wu, C.~Winter, C.~Hesse, M.~Chen, E.~Sigler, M.~Litwin, S.~Gray, B.~Chess, J.~Clark, C.~Berner, S.~McCandlish, A.~Radford, I.~Sutskever, D.~Amodei, \href{https://proceedings.neurips.cc/paper/2020/hash/1457c0d6bfcb4967418bfb8ac142f64a-Abstract.html}{Language models are few-shot learners}, in: H.~Larochelle, M.~Ranzato, R.~Hadsell, M.~Balcan, H.~Lin (Eds.), Advances in Neural Information Processing Systems 33: Annual Conference on Neural Information Processing Systems 2020, NeurIPS 2020, December 6-12, 2020, virtual, 2020.
\newline\urlprefix\url{https://proceedings.neurips.cc/paper/2020/hash/1457c0d6bfcb4967418bfb8ac142f64a-Abstract.html}

\bibitem{Radford2019}
A.~Radford, J.~Wu, R.~Child, D.~Luan, D.~Amodei, I.~Sutskever, Language models are unsupervised multitask learners, OpenAI Blog (2019).

\bibitem{flan}
J.~Wei, M.~Bosma, V.~Y. Zhao, K.~Guu, A.~W. Yu, B.~Lester, N.~Du, A.~M. Dai, Q.~V. Le, \href{https://arxiv.org/abs/2109.01652}{Finetuned language models are zero-shot learners}, CoRR abs/2109.01652 (2021).
\newblock \href {http://arxiv.org/abs/2109.01652} {\path{arXiv:2109.01652}}.
\newline\urlprefix\url{https://arxiv.org/abs/2109.01652}

\bibitem{roberta}
Y.~Liu, M.~Ott, N.~Goyal, J.~Du, M.~Joshi, D.~Chen, O.~Levy, M.~Lewis, L.~Zettlemoyer, V.~Stoyanov, \href{http://arxiv.org/abs/1907.11692}{Roberta: {A} robustly optimized {BERT} pretraining approach}, CoRR abs/1907.11692 (2019).
\newblock \href {http://arxiv.org/abs/1907.11692} {\path{arXiv:1907.11692}}.
\newline\urlprefix\url{http://arxiv.org/abs/1907.11692}

\bibitem{debertav3}
P.~He, J.~Gao, W.~Chen, \href{https://openreview.net/forum?id=sE7-XhLxHA}{Debertav3: Improving deberta using electra-style pre-training with gradient-disentangled embedding sharing}, in: The Eleventh International Conference on Learning Representations, {ICLR} 2023, Kigali, Rwanda, May 1-5, 2023, OpenReview.net, 2023.
\newline\urlprefix\url{https://openreview.net/forum?id=sE7-XhLxHA}

\bibitem{modernbertbp}
B.~Warner, A.~Chaffin, B.~Clavié, O.~Weller, O.~Hallström, S.~Taghadouini, A.~Gallagher, R.~Biswas, F.~Ladhak, T.~Aarsen, N.~Cooper, G.~Adams, J.~Howard, J.~Whitaker, I.~Poli, \href{https://huggingface.co/blog/modernbert}{Finally, a replacement for bert: Introducing modernbert}, hugging Face Blog (December 2024).
\newline\urlprefix\url{https://huggingface.co/blog/modernbert}

\bibitem{cot}
J.~Wei, X.~Wang, D.~Schuurmans, M.~Bosma, E.~H. Chi, Q.~Le, D.~Zhou, \href{https://arxiv.org/abs/2201.11903}{Chain of thought prompting elicits reasoning in large language models}, CoRR abs/2201.11903 (2022).
\newblock \href {http://arxiv.org/abs/2201.11903} {\path{arXiv:2201.11903}}.
\newline\urlprefix\url{https://arxiv.org/abs/2201.11903}

\bibitem{clfsurvey}
Q.~Li, H.~Peng, J.~Li, C.~Xia, R.~Yang, L.~Sun, P.~S. Yu, L.~He, \href{https://doi.org/10.1145/3495162}{A survey on text classification: From traditional to deep learning}, ACM Trans. Intell. Syst. Technol. 13~(2) (Apr. 2022).
\newblock \href {https://doi.org/10.1145/3495162} {\path{doi:10.1145/3495162}}.
\newline\urlprefix\url{https://doi.org/10.1145/3495162}

\bibitem{MNLI}
A.~Williams, N.~Nangia, S.~Bowman, A broad-coverage challenge corpus for sentence understanding through inference, in: Proceedings of the 2018 Conference of the North American Chapter of the Association for Computational Linguistics: Human Language Technologies, Volume 1 (Long Papers), 2018, pp. 1112--1122.

\bibitem{bart}
M.~Lewis, Y.~Liu, N.~Goyal, M.~Ghazvininejad, A.~Mohamed, O.~Levy, V.~Stoyanov, L.~Zettlemoyer, \href{https://aclanthology.org/2020.acl-main.703/}{{BART}: Denoising sequence-to-sequence pre-training for natural language generation, translation, and comprehension}, in: D.~Jurafsky, J.~Chai, N.~Schluter, J.~Tetreault (Eds.), Proceedings of the 58th Annual Meeting of the Association for Computational Linguistics, Association for Computational Linguistics, Online, 2020, pp. 7871--7880.
\newblock \href {https://doi.org/10.18653/v1/2020.acl-main.703} {\path{doi:10.18653/v1/2020.acl-main.703}}.
\newline\urlprefix\url{https://aclanthology.org/2020.acl-main.703/}

\bibitem{tasksource}
D.~Sileo, \href{https://aclanthology.org/2024.lrec-main.1361}{tasksource: A large collection of {NLP} tasks with a structured dataset preprocessing framework}, in: Proceedings of the 2024 Joint International Conference on Computational Linguistics, Language Resources and Evaluation (LREC-COLING 2024), ELRA and ICCL, Torino, Italia, 2024, pp. 15655--15684.
\newline\urlprefix\url{https://aclanthology.org/2024.lrec-main.1361}

\bibitem{sst}
C.~Liu, W.~Zhang, G.~Chen, X.~Wu, A.~T. Luu, C.~H. Chang, L.~Bing, \href{https://aclanthology.org/2023.findings-acl.110/}{Zero-shot text classification via self-supervised tuning}, in: A.~Rogers, J.~Boyd-Graber, N.~Okazaki (Eds.), Findings of the Association for Computational Linguistics: ACL 2023, Association for Computational Linguistics, Toronto, Canada, 2023, pp. 1743--1761.
\newblock \href {https://doi.org/10.18653/v1/2023.findings-acl.110} {\path{doi:10.18653/v1/2023.findings-acl.110}}.
\newline\urlprefix\url{https://aclanthology.org/2023.findings-acl.110/}

\bibitem{cloze}
W.~L. Taylor, “cloze procedure”: A new tool for measuring readability, Journalism quarterly 30~(4) (1953) 415--433.

\bibitem{PET}
T.~Schick, H.~Sch{\"{u}}tze, \href{https://doi.org/10.18653/v1/2021.eacl-main.20}{Exploiting cloze-questions for few-shot text classification and natural language inference}, in: P.~Merlo, J.~Tiedemann, R.~Tsarfaty (Eds.), Proceedings of the 16th Conference of the European Chapter of the Association for Computational Linguistics: Main Volume, {EACL} 2021, Online, April 19 - 23, 2021, Association for Computational Linguistics, 2021, pp. 255--269.
\newblock \href {https://doi.org/10.18653/V1/2021.EACL-MAIN.20} {\path{doi:10.18653/V1/2021.EACL-MAIN.20}}.
\newline\urlprefix\url{https://doi.org/10.18653/v1/2021.eacl-main.20}

\bibitem{petfewshot}
T.~Schick, H.~Sch{\"u}tze, \href{https://aclanthology.org/2021.naacl-main.185/}{It`s not just size that matters: Small language models are also few-shot learners}, in: K.~Toutanova, A.~Rumshisky, L.~Zettlemoyer, D.~Hakkani-Tur, I.~Beltagy, S.~Bethard, R.~Cotterell, T.~Chakraborty, Y.~Zhou (Eds.), Proceedings of the 2021 Conference of the North American Chapter of the Association for Computational Linguistics: Human Language Technologies, Association for Computational Linguistics, Online, 2021, pp. 2339--2352.
\newblock \href {https://doi.org/10.18653/v1/2021.naacl-main.185} {\path{doi:10.18653/v1/2021.naacl-main.185}}.
\newline\urlprefix\url{https://aclanthology.org/2021.naacl-main.185/}

\bibitem{bertgenerative}
D.~Samuel, \href{https://doi.org/10.48550/arXiv.2406.04823}{Berts are generative in-context learners}, CoRR abs/2406.04823 (2024).
\newblock \href {http://arxiv.org/abs/2406.04823} {\path{arXiv:2406.04823}}, \href {https://doi.org/10.48550/ARXIV.2406.04823} {\path{doi:10.48550/ARXIV.2406.04823}}.
\newline\urlprefix\url{https://doi.org/10.48550/arXiv.2406.04823}

\bibitem{unimc}
P.~Yang, J.~Wang, R.~Gan, X.~Zhu, L.~Zhang, Z.~Wu, X.~Gao, J.~Zhang, T.~Sakai, \href{https://aclanthology.org/2022.emnlp-main.474/}{Zero-shot learners for natural language understanding via a unified multiple choice perspective}, in: Y.~Goldberg, Z.~Kozareva, Y.~Zhang (Eds.), Proceedings of the 2022 Conference on Empirical Methods in Natural Language Processing, Association for Computational Linguistics, Abu Dhabi, United Arab Emirates, 2022, pp. 7042--7055.
\newblock \href {https://doi.org/10.18653/v1/2022.emnlp-main.474} {\path{doi:10.18653/v1/2022.emnlp-main.474}}.
\newline\urlprefix\url{https://aclanthology.org/2022.emnlp-main.474/}

\bibitem{deepseek}
A.~Liu, B.~Feng, B.~Xue, B.~Wang, B.~Wu, C.~Lu, C.~Zhao, C.~Deng, C.~Zhang, C.~Ruan, et~al., Deepseek-v3 technical report, arXiv preprint arXiv:2412.19437 (2024).

\bibitem{llama2}
H.~Touvron, L.~Martin, K.~Stone, P.~Albert, A.~Almahairi, Y.~Babaei, N.~Bashlykov, S.~Batra, P.~Bhargava, S.~Bhosale, et~al., Llama 2: Open foundation and fine-tuned chat models, arXiv preprint arXiv:2307.09288 (2023).

\bibitem{qwen2}
A.~Yang, B.~Yang, B.~Hui, B.~Zheng, B.~Yu, C.~Zhou, C.~Li, C.~Li, D.~Liu, F.~Huang, et~al., Qwen2 technical report, arXiv preprint arXiv:2407.10671 (2024).

\bibitem{dolma}
L.~Soldaini, R.~Kinney, A.~Bhagia, D.~Schwenk, D.~Atkinson, R.~Authur, B.~Bogin, K.~Chandu, J.~Dumas, Y.~Elazar, et~al., Dolma: An open corpus of three trillion tokens for language model pretraining research, arXiv preprint arXiv:2402.00159 (2024).

\bibitem{dclm}
J.~Li, A.~Fang, G.~Smyrnis, M.~Ivgi, M.~Jordan, S.~Gadre, H.~Bansal, E.~Guha, S.~Keh, K.~Arora, et~al., Datacomp-lm: In search of the next generation of training sets for language models, arXiv preprint arXiv:2406.11794 (2024).

\bibitem{fineweb}
G.~Penedo, H.~Kydl{\'{\i}}cek, L.~B. Allal, A.~Lozhkov, M.~Mitchell, C.~Raffel, L.~von Werra, T.~Wolf, \href{https://doi.org/10.48550/arXiv.2406.17557}{The fineweb datasets: Decanting the web for the finest text data at scale}, CoRR abs/2406.17557 (2024).
\newblock \href {http://arxiv.org/abs/2406.17557} {\path{arXiv:2406.17557}}, \href {https://doi.org/10.48550/ARXIV.2406.17557} {\path{doi:10.48550/ARXIV.2406.17557}}.
\newline\urlprefix\url{https://doi.org/10.48550/arXiv.2406.17557}

\bibitem{modernbert}
B.~Warner, A.~Chaffin, B.~Clavié, O.~Weller, O.~Hallström, S.~Taghadouini, A.~Gallagher, R.~Biswas, F.~Ladhak, T.~Aarsen, N.~Cooper, G.~Adams, J.~Howard, I.~Poli, \href{https://arxiv.org/abs/2412.13663}{Smarter, better, faster, longer: A modern bidirectional encoder for fast, memory efficient, and long context finetuning and inference} (2024).
\newblock \href {http://arxiv.org/abs/2412.13663} {\path{arXiv:2412.13663}}.
\newline\urlprefix\url{https://arxiv.org/abs/2412.13663}

\bibitem{smollm2}
L.~B. Allal, A.~Lozhkov, E.~Bakouch, G.~M. Blázquez, L.~Tunstall, A.~Piqueres, A.~Marafioti, C.~Zakka, L.~von Werra, T.~Wolf, Smollm2 - with great data, comes great performance (2024).

\bibitem{gte}
X.~Zhang, Y.~Zhang, D.~Long, W.~Xie, Z.~Dai, J.~Tang, H.~Lin, B.~Yang, P.~Xie, F.~Huang, M.~Zhang, W.~Li, M.~Zhang, \href{https://aclanthology.org/2024.emnlp-industry.103}{mgte: Generalized long-context text representation and reranking models for multilingual text retrieval}, in: F.~Dernoncourt, D.~Preotiuc{-}Pietro, A.~Shimorina (Eds.), Proceedings of the 2024 Conference on Empirical Methods in Natural Language Processing: {EMNLP} 2024 - Industry Track, Miami, Florida, USA, November 12-16, 2024, Association for Computational Linguistics, 2024, pp. 1393--1412.
\newline\urlprefix\url{https://aclanthology.org/2024.emnlp-industry.103}

\bibitem{phi}
M.~Javaheripi, S.~Bubeck, M.~Abdin, J.~Aneja, S.~Bubeck, C.~C.~T. Mendes, W.~Chen, A.~Del~Giorno, R.~Eldan, S.~Gopi, et~al., Phi-2: The surprising power of small language models, Microsoft Research Blog 1~(3) (2023) 3.

\bibitem{mamba}
A.~Gu, T.~Dao, \href{https://doi.org/10.48550/arXiv.2312.00752}{Mamba: Linear-time sequence modeling with selective state spaces}, CoRR abs/2312.00752 (2023).
\newblock \href {http://arxiv.org/abs/2312.00752} {\path{arXiv:2312.00752}}, \href {https://doi.org/10.48550/ARXIV.2312.00752} {\path{doi:10.48550/ARXIV.2312.00752}}.
\newline\urlprefix\url{https://doi.org/10.48550/arXiv.2312.00752}

\bibitem{flant5}
H.~W. Chung, L.~Hou, S.~Longpre, B.~Zoph, Y.~Tay, W.~Fedus, Y.~Li, X.~Wang, M.~Dehghani, S.~Brahma, et~al., Scaling instruction-finetuned language models, Journal of Machine Learning Research 25~(70) (2024) 1--53.

\bibitem{deberta}
P.~He, J.~Gao, W.~Chen, \href{https://arxiv.org/abs/2111.09543}{Debertav3: Improving deberta using electra-style pre-training with gradient-disentangled embedding sharing} (2023).
\newblock \href {http://arxiv.org/abs/2111.09543} {\path{arXiv:2111.09543}}.
\newline\urlprefix\url{https://arxiv.org/abs/2111.09543}

\bibitem{seq2seq}
I.~Sutskever, O.~Vinyals, Q.~V. Le, \href{https://proceedings.neurips.cc/paper/2014/hash/a14ac55a4f27472c5d894ec1c3c743d2-Abstract.html}{Sequence to sequence learning with neural networks}, in: Z.~Ghahramani, M.~Welling, C.~Cortes, N.~D. Lawrence, K.~Q. Weinberger (Eds.), Advances in Neural Information Processing Systems 27: Annual Conference on Neural Information Processing Systems 2014, December 8-13 2014, Montreal, Quebec, Canada, 2014, pp. 3104--3112.
\newline\urlprefix\url{https://proceedings.neurips.cc/paper/2014/hash/a14ac55a4f27472c5d894ec1c3c743d2-Abstract.html}

\bibitem{t5}
C.~Raffel, N.~Shazeer, A.~Roberts, K.~Lee, S.~Narang, M.~Matena, Y.~Zhou, W.~Li, P.~J. Liu, Exploring the limits of transfer learning with a unified text-to-text transformer, Journal of machine learning research 21~(140) (2020) 1--67.

\bibitem{promptelectra}
M.~Xia, M.~Artetxe, J.~Du, D.~Chen, V.~Stoyanov, \href{https://aclanthology.org/2022.emnlp-main.780/}{Prompting {ELECTRA}: Few-shot learning with discriminative pre-trained models}, in: Y.~Goldberg, Z.~Kozareva, Y.~Zhang (Eds.), Proceedings of the 2022 Conference on Empirical Methods in Natural Language Processing, Association for Computational Linguistics, Abu Dhabi, United Arab Emirates, 2022, pp. 11351--11361.
\newblock \href {https://doi.org/10.18653/v1/2022.emnlp-main.780} {\path{doi:10.18653/v1/2022.emnlp-main.780}}.
\newline\urlprefix\url{https://aclanthology.org/2022.emnlp-main.780/}

\bibitem{flancol}
S.~Longpre, L.~Hou, T.~Vu, A.~Webson, H.~W. Chung, Y.~Tay, D.~Zhou, Q.~V. Le, B.~Zoph, J.~Wei, A.~Roberts, \href{https://proceedings.mlr.press/v202/longpre23a.html}{The flan collection: Designing data and methods for effective instruction tuning}, in: A.~Krause, E.~Brunskill, K.~Cho, B.~Engelhardt, S.~Sabato, J.~Scarlett (Eds.), International Conference on Machine Learning, {ICML} 2023, 23-29 July 2023, Honolulu, Hawaii, {USA}, Vol. 202 of Proceedings of Machine Learning Research, {PMLR}, 2023, pp. 22631--22648.
\newline\urlprefix\url{https://proceedings.mlr.press/v202/longpre23a.html}

\bibitem{mmlu}
D.~Hendrycks, C.~Burns, S.~Basart, A.~Zou, M.~Mazeika, D.~Song, J.~Steinhardt, Measuring massive multitask language understanding, Proceedings of the International Conference on Learning Representations (ICLR) (2021).

\bibitem{bbh}
M.~Suzgun, N.~Scales, N.~Sch{\"a}rli, S.~Gehrmann, Y.~Tay, H.~W. Chung, A.~Chowdhery, Q.~V. Le, E.~H. Chi, D.~Zhou, , J.~Wei, Challenging big-bench tasks and whether chain-of-thought can solve them, arXiv preprint arXiv:2210.09261 (2022).

\bibitem{dropout}
N.~Srivastava, G.~Hinton, A.~Krizhevsky, I.~Sutskever, R.~Salakhutdinov, Dropout: a simple way to prevent neural networks from overfitting, The journal of machine learning research 15~(1) (2014) 1929--1958.

\bibitem{colbert}
O.~Khattab, M.~Zaharia, Colbert: Efficient and effective passage search via contextualized late interaction over bert, in: Proceedings of the 43rd International ACM SIGIR conference on research and development in Information Retrieval, 2020, pp. 39--48.

\bibitem{e5}
L.~Wang, N.~Yang, X.~Huang, B.~Jiao, L.~Yang, D.~Jiang, R.~Majumder, F.~Wei, Text embeddings by weakly-supervised contrastive pre-training, arXiv preprint arXiv:2212.03533 (2022).

\bibitem{stella}
D.~Zhang, et~al., Jasper and stella: distillation of sota embedding models, arXiv e-prints (2024) arXiv--2412.

\bibitem{mmlupro}
Y.~Wang, X.~Ma, G.~Zhang, Y.~Ni, A.~Chandra, S.~Guo, W.~Ren, A.~Arulraj, X.~He, Z.~Jiang, et~al., Mmlu-pro: A more robust and challenging multi-task language understanding benchmark, arXiv preprint arXiv:2406.01574 (2024).

\bibitem{raft}
N.~Alex, E.~Lifland, L.~Tunstall, A.~Thakur, P.~Maham, C.~J. Riedel, E.~Hine, C.~Ashurst, P.~Sedille, A.~Carlier, M.~Noetel, A.~Stuhlm{\"u}ller, \href{https://api.semanticscholar.org/CorpusID:238215290}{Raft: A real-world few-shot text classification benchmark}, ArXiv abs/2109.14076 (2021).
\newline\urlprefix\url{https://api.semanticscholar.org/CorpusID:238215290}

\bibitem{ade}
H.~Gurulingappa, A.~M. Rajput, A.~Roberts, J.~Fluck, M.~Hofmann-Apitius, L.~Toldo, \href{https://www.sciencedirect.com/science/article/pii/S1532046412000615}{Development of a benchmark corpus to support the automatic extraction of drug-related adverse effects from medical case reports}, Journal of Biomedical Informatics 45~(5) (2012) 885--892, text Mining and Natural Language Processing in Pharmacogenomics.
\newblock \href {https://doi.org/https://doi.org/10.1016/j.jbi.2012.04.008} {\path{doi:https://doi.org/10.1016/j.jbi.2012.04.008}}.
\newline\urlprefix\url{https://www.sciencedirect.com/science/article/pii/S1532046412000615}

\bibitem{onestopenglish}
S.~Vajjala, I.~Lu{\v{c}}i{\'c}, \href{https://aclanthology.org/W18-0535/}{{O}ne{S}top{E}nglish corpus: A new corpus for automatic readability assessment and text simplification}, in: J.~Tetreault, J.~Burstein, E.~Kochmar, C.~Leacock, H.~Yannakoudakis (Eds.), Proceedings of the Thirteenth Workshop on Innovative Use of {NLP} for Building Educational Applications, Association for Computational Linguistics, New Orleans, Louisiana, 2018, pp. 297--304.
\newblock \href {https://doi.org/10.18653/v1/W18-0535} {\path{doi:10.18653/v1/W18-0535}}.
\newline\urlprefix\url{https://aclanthology.org/W18-0535/}

\bibitem{lmharness}
L.~Gao, J.~Tow, B.~Abbasi, S.~Biderman, S.~Black, A.~DiPofi, C.~Foster, L.~Golding, J.~Hsu, A.~Le~Noac'h, H.~Li, K.~McDonell, N.~Muennighoff, C.~Ociepa, J.~Phang, L.~Reynolds, H.~Schoelkopf, A.~Skowron, L.~Sutawika, E.~Tang, A.~Thite, B.~Wang, K.~Wang, A.~Zou, \href{https://zenodo.org/records/12608602}{A framework for few-shot language model evaluation} (07 2024).
\newblock \href {https://doi.org/10.5281/zenodo.12608602} {\path{doi:10.5281/zenodo.12608602}}.
\newline\urlprefix\url{https://zenodo.org/records/12608602}

\bibitem{yahoo}
X.~Zhang, J.~Zhao, Y.~LeCun, Character-level convolutional networks for text classification, Advances in neural information processing systems 28 (2015).

\bibitem{imdb}
A.~L. Maas, R.~E. Daly, P.~T. Pham, D.~Huang, A.~Y. Ng, C.~Potts, \href{https://aclanthology.org/P11-1015/}{Learning word vectors for sentiment analysis}, in: D.~Lin, Y.~Matsumoto, R.~Mihalcea (Eds.), Proceedings of the 49th Annual Meeting of the Association for Computational Linguistics: Human Language Technologies, Association for Computational Linguistics, Portland, Oregon, USA, 2011, pp. 142--150.
\newline\urlprefix\url{https://aclanthology.org/P11-1015/}

\bibitem{sst2}
Y.~Liu, M.~Ott, N.~Goyal, J.~Du, M.~Joshi, D.~Chen, O.~Levy, M.~Lewis, L.~Zettlemoyer, V.~Stoyanov, \href{http://arxiv.org/abs/1907.11692}{Roberta: {A} robustly optimized {BERT} pretraining approach}, CoRR abs/1907.11692 (2019).
\newblock \href {http://arxiv.org/abs/1907.11692} {\path{arXiv:1907.11692}}.
\newline\urlprefix\url{http://arxiv.org/abs/1907.11692}

\bibitem{wang-etal-2018-glue}
A.~Wang, A.~Singh, J.~Michael, F.~Hill, O.~Levy, S.~Bowman, \href{https://aclanthology.org/W18-5446}{{GLUE}: A multi-task benchmark and analysis platform for natural language understanding}, in: T.~Linzen, G.~Chrupa{\l}a, A.~Alishahi (Eds.), Proceedings of the 2018 {EMNLP} Workshop {B}lackbox{NLP}: Analyzing and Interpreting Neural Networks for {NLP}, Association for Computational Linguistics, Brussels, Belgium, 2018, pp. 353--355.
\newblock \href {https://doi.org/10.18653/v1/W18-5446} {\path{doi:10.18653/v1/W18-5446}}.
\newline\urlprefix\url{https://aclanthology.org/W18-5446}

\bibitem{openorca}
W.~Lian, B.~Goodson, E.~Pentland, A.~Cook, C.~Vong, "Teknium", Openorca: An open dataset of gpt augmented flan reasoning traces, \url{https://https://huggingface.co/Open-Orca/OpenOrca} (2023).

\bibitem{hermes}
R.~Teknium, J.~Quesnelle, C.~Guang, \href{https://arxiv.org/abs/2408.11857}{Hermes 3 technical report} (2024).
\newblock \href {http://arxiv.org/abs/2408.11857} {\path{arXiv:2408.11857}}.
\newline\urlprefix\url{https://arxiv.org/abs/2408.11857}

\bibitem{icl}
Q.~Dong, L.~Li, D.~Dai, C.~Zheng, J.~Ma, R.~Li, H.~Xia, J.~Xu, Z.~Wu, T.~Liu, B.~Chang, X.~Sun, L.~Li, Z.~Sui, \href{https://arxiv.org/abs/2301.00234}{A survey on in-context learning} (2024).
\newblock \href {http://arxiv.org/abs/2301.00234} {\path{arXiv:2301.00234}}.
\newline\urlprefix\url{https://arxiv.org/abs/2301.00234}

\bibitem{ul2}
Y.~Tay, M.~Dehghani, V.~Q. Tran, X.~Garcia, J.~Wei, X.~Wang, H.~W. Chung, D.~Bahri, T.~Schuster, H.~S. Zheng, D.~Zhou, N.~Houlsby, D.~Metzler, \href{https://openreview.net/forum?id=6ruVLB727MC}{{UL2:} unifying language learning paradigms}, in: The Eleventh International Conference on Learning Representations, {ICLR} 2023, Kigali, Rwanda, May 1-5, 2023, OpenReview.net, 2023.
\newline\urlprefix\url{https://openreview.net/forum?id=6ruVLB727MC}

\end{thebibliography}






\end{document}